\documentclass{article}

\PassOptionsToPackage{numbers, compress}{natbib}

    \usepackage[final]{neurips_2025}

\usepackage[utf8]{inputenc} 
\usepackage[T1]{fontenc}    
\usepackage{hyperref}       
\usepackage{url}            
\usepackage{booktabs}       
\usepackage{amsfonts}       
\usepackage{nicefrac}       
\usepackage{microtype}      
\usepackage[table]{xcolor}
\usepackage{color}
\usepackage{amsmath}
\usepackage{graphicx}
\usepackage{multirow}
\usepackage{float,caption}
\usepackage{wrapfig}
\usepackage{pifont}

\title{GaussianFusion: Gaussian-Based Multi-Sensor Fusion for End-to-End Autonomous Driving}

\author{
Shuai~Liu,~Quanmin~Liang,~Zefeng~Li,~Boyang~Li\thanks{Corresponding author.},~Kai~Huang \\
School of Computer Science and Engineering, Sun Yat-sen University\\
\{liush376@mail2, liby83@mail, huangk36@mail\}.sysu.edu.cn
}

\begin{document}

\maketitle

\begin{abstract}
Multi-sensor fusion is crucial for improving the performance and robustness of end-to-end autonomous driving systems. Existing methods predominantly adopt either attention-based flatten fusion or bird’s eye view fusion through geometric transformations. However, these approaches often suffer from limited interpretability or dense computational overhead. In this paper, we introduce \textbf{GaussianFusion}, a \textbf{Gaussian}-based multi-sensor \textbf{fusion} framework for end-to-end autonomous driving. Our method employs explicit and compact Gaussian representations as intermediate carriers to aggregate information from diverse sensors. Specifically, we initialize a set of 2D Gaussians uniformly across the driving scene, where each Gaussian is parameterized by physical attributes and equipped with explicit and implicit features. These Gaussians are progressively refined by integrating multi-modal features. The explicit features capture rich semantic and spatial information about the traffic scene, while the implicit features provide complementary cues beneficial for trajectory planning. To fully exploit rich spatial and semantic information in Gaussians, we design a cascade planning head that iteratively refines trajectory predictions through interactions with Gaussians. Extensive experiments on the NAVSIM and Bench2Drive benchmarks demonstrate the effectiveness and robustness of the proposed GaussianFusion framework. The source code will be released at \url{https://github.com/Say2L/GaussianFusion}. 

\end{abstract}

\section{Introduction}

End-to-end (E2E) autonomous driving \cite{uniad, vad, vadv2, transfuser, diffusiondrive} has attracted growing attention for its potential to streamline the traditional modular pipeline by directly mapping sensor inputs to driving actions through deep learning. This paradigm reduces system complexity and enables joint optimization across tasks. However, relying on a single sensor often limits the system’s ability to handle diverse and challenging driving scenarios. To address this limitation, multi-sensor fusion has become essential, as it allows the model to leverage complementary information from different sensors such as cameras, LiDARs, and radars. This integration enhances perception reliability and provides richer input for learning robust driving policies.

\begin{wrapfigure}{r}{0.5\textwidth} 
  \centering
  \includegraphics[width=0.48\textwidth]{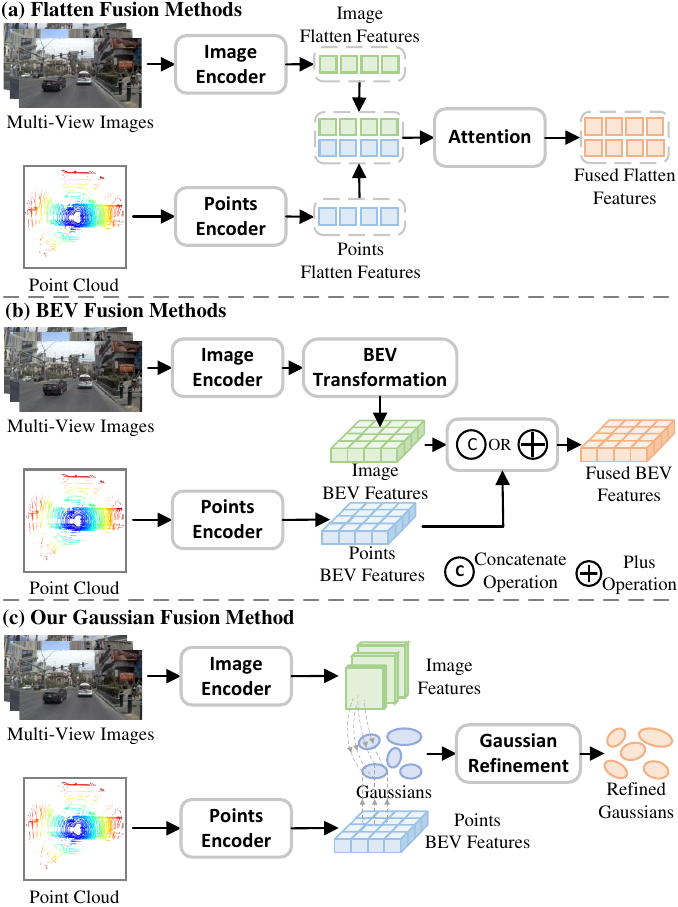} 
  \caption{\textbf{Pipelines of different multi-sensor fusion methods.} }
  \label{fig_fusion}
\end{wrapfigure}

Existing multi-modal fusion strategies in E2E autonomous driving can be broadly divided into two categories: flatten fusion and bird’s eye view (BEV) fusion. Flatten fusion methods \cite{transfuser, interfuser, bai2022transfusion, autoalign} typically compress sensor features, such as those from images and LiDAR point clouds, into a shared latent space, where feature interaction is performed using attention mechanisms, as shown in Figure~\ref{fig_fusion} (a). These approaches are appealing due to their flexibility and efficiency, often requiring minimal geometric calibration. However, the lack of explicit spatial grounding in 3D space limits their interpretability and makes them less effective in scenarios requiring precise spatial reasoning. In contrast, BEV fusion methods \cite{thinktwice, lss, li2024bevformer, fusionad} project multi-modal features into a common BEV coordinate, leveraging geometric priors to align data from different sensors, as shown in Figure~\ref{fig_fusion} (b). This facilitates structured spatial understanding and improves performance in downstream perception tasks such as 3D object detection and map construction. Nevertheless, BEV fusion incurs significant computational and memory overhead due to the dense nature of the BEV representation, especially when high-resolution inputs or fine-grained features are involved. As a result, there remains an ongoing challenge to develop fusion frameworks that can balance spatial awareness, efficiency, and scalability in complex driving environments.

3D Gaussians have recently gained attention in camera-based 3D scene representation and reconstruction \cite{gss, huang2024gaussianformer, gaussianformerv2} due to their physical interpretability, compactness, and inherent sparsity. These properties make them promising candidates for multi-sensor fusion in autonomous driving, where efficiency and structured spatial understanding are critical. However, applying Gaussian representations in this context introduces several challenges. First, it is difficult to effectively supervise Gaussian parameters due to the lack of fine-grained 3D scene annotations in existing E2E driving datasets \cite{navsim, bench2drive, hidden_datasets}. Second, existing approaches \cite{huang2024gaussianformer, gaussianformerv2} mainly focus on 3D scene representation, leaving their suitability for the motion planning task underexplored. Third, efficiently leveraging Gaussian representation for accurate trajectory generation requires careful architecture design. Addressing these issues is key to enabling Gaussian-based representations in E2E autonomous driving frameworks.

With the aforementioned innovations and considerations, we propose GaussianFusion, a Gaussian-based multi-sensor fusion framework for E2E autonomous driving. Our approach leverages 2D Gaussians to represent the traffic scene, offering improved efficiency over 3D Gaussians. Notably, 2D Gaussians require supervision only from BEV semantic maps, which are widely available in E2E datasets \cite{navsim, bench2drive, hidden_datasets}. To tailor the fusion process to the motion planning task, we design a dual-branch fusion pipeline. The first branch captures local features from multi-sensor inputs for each Gaussian and is primarily responsible for traffic scene reconstruction. The second branch aggregates global planning cues from the same inputs and is dedicated exclusively to motion planning. Moreover, to fully exploit the representational capacity of Gaussians, we introduce a cascade planning module that refines the anchor trajectories by querying the Gaussian representations in a cascaded manner.

We evaluate GaussianFusion on the planning-oriented NAVSIM dataset \cite{navsim}. Utilizing the V2-99 backbone \cite{V299}, our approach achieves 92.0 PDMS \cite{navsim}, significantly surpassing current state-of-the-art methods. To further evaluate the generalization and robustness of our framework, we conduct experiments on the closed-loop benchmark Bench2Drive \cite{bench2drive}, where the results consistently demonstrate the effectiveness of GaussianFusion. The key contributions of this work are summarized as follows:

\begin{itemize}
\item We first introduce Gaussian representations into the domain of multi-sensor fusion for E2E autonomous driving, and we propose a dual-branch fusion pipeline tailored to the planning-centric task.
\item We design a cascade planning head specifically adapted to the Gaussian representation, which iteratively refines trajectories through hierarchical Gaussian queries.
\item Extensive evaluations on both open-loop (NAVSIM) and closed-loop (Bench2Drive) benchmarks demonstrate the superior performance and robustness of GaussianFusion.
\end{itemize}

\section{Related Work}
\noindent\textbf{End-to-End Autonomous Driving. }
E2E autonomous driving approaches can be broadly categorized into single-task and multi-task methods. Single-task methods \cite{2016end, 2018end, 2019exploring} primarily focus on the planning task. An early work \cite{2016end} employs a convolutional neural network (CNN) to directly map raw pixel data from a front-facing camera to steering commands. Building on this idea, \cite{2018end} proposes conditional imitation learning, which extends traditional imitation learning by integrating high-level navigational commands to better capture driver intent. CILRS \cite{2019exploring} further explores behavior cloning within the E2E framework, aiming to improve performance through more structured supervision.

Multi-task methods extend the E2E autonomous driving paradigm by incorporating auxiliary tasks such as 3D object detection, mapping, and motion forecasting. Compared to single-task approaches, the multi-task paradigm enables the learning of scene representations with improved generalization and interpretability \cite{2021multi}. This has led to growing interest from the research community. UniAD \cite{uniad} introduces a planning-oriented framework that sequentially integrates perception, prediction, and planning modules. VAD \cite{vad} and its improved version VADv2 \cite{vadv2} adopt a vectorized scene representation to enhance computational efficiency. More recently, diffusion-based strategies \cite{chi2023diffusion} have gained attention in E2E autonomous driving. GoalFlow \cite{goalflow} introduces goal-conditioned denoising, leveraging goal points to guide the generation of multi-modal trajectories. DiffusionDrive \cite{diffusiondrive} proposes a truncated diffusion policy to accelerate the denoising process, improving inference speed without sacrificing trajectory quality.


\noindent\textbf{Sensor Fusion for Autonomous Driving. }
Existing sensor fusion approaches can be generally classified into flatten fusion and BEV fusion methods. Flatten fusion methods \cite{interfuser, transfuser, tf++, bai2022transfusion, autoalign} typically compress image and LiDAR features into flattened representations, which are then fused using attention mechanisms \cite{attention, cross-attention}. For example, InterFuser \cite{interfuser} concatenates flattened features from multiple sensors and applies self-attention \cite{attention} to enable information exchange. TransFuser \cite{transfuser, tf++} follows the same paradigm but incorporates multi-scale features to enhance fusion. To mitigate the influence of irrelevant image regions, TransFusion \cite{bai2022transfusion} introduces a 2D circular Gaussian mask that modulates the cross-attention weights. AutoAlign \cite{autoalign} further improves upon this by introducing a cross-attention-based feature alignment module that adaptively aggregates pixel-level image features for each voxel. While these methods eliminate the need for explicit calibration, they often suffer from limited interpretability and lack the capability for explicit 3D scene modeling.

BEV fusion methods \cite{liang2018deep, thinktwice, lss, li2024bevformer, fusionad} typically fuse multi-modal features within the BEV plane based on geometric projection relationships. MILE \cite{hu2022model} lifts image features into 3D using predicted depth distributions, then aggregates voxel features into BEV via grid-based sum-pooling. ContFuse \cite{liang2018deep} employs 3D point clouds as intermediaries to project pixel-level image features onto the BEV plane for fusion. ThinkTwice \cite{thinktwice} and DriveAdapter \cite{driveadapter} adopt the LSS framework \cite{lss} to transform image features into BEV space, which are then concatenated with LiDAR-derived BEV features. BEVFormer \cite{li2024bevformer} and FusionAD \cite{fusionad} further extend the BEV representation by incorporating a height dimension into the BEV queries, enabling spatially-aware feature extraction from image inputs. This type of method addresses the interpretability and 3D modeling limitations inherent in flatten fusion methods. However, since operating on a dense BEV grid, these methods often incur significant computational and memory overhead.


In contrast to existing fusion paradigms, we propose GaussianFusion, which leverages Gaussian representations as intermediaries to extract and aggregate multi-modal features. Gaussians are well-regarded for their strong scene reconstruction \cite{gss} and explicit representation properties. Moreover, Gaussian representations are sparsely distributed in space, in stark contrast to the dense grids used in BEV fusion methods. As a result, our GaussianFusion effectively circumvents the limitations of both flatten fusion (lack of structure and interpretability) and BEV fusion (computational and memory inefficiency), offering a more efficient and interpretable alternative for multi-sensor integration.

\noindent\textbf{Gaussian Representation. }
3D Gaussian Splatting (3D-GS) \cite{gss} first introduced 3D Gaussians as a scene representation for radiance field rendering, achieving high visual fidelity and real-time performance. Deformable-GS \cite{yang2024deformable} extends this technique to dynamic scene reconstruction and rendering. Other works \cite{tang2023dreamgaussian, yi2023gaussiandreamer} adapt Gaussian splatting for 3D content creation. The GaussianFormer series \cite{huang2024gaussianformer, gaussianformerv2} utilizes 3D Gaussians to perform occupancy prediction. 

The approach most closely related to ours is GaussianAD \cite{gaussianad}, which models future scene evolution using Gaussian flows. While both methods leverage Gaussians for representing traffic scenes, our GaussianFusion differs from GaussianAD in several key aspects: 1) GaussianFusion is designed for multi-sensor fusion in E2E autonomous driving, whereas GaussianAD focuses solely on vision-based systems. 2) Our method employs 2D Gaussians as scene representations, in contrast to the 3D Gaussians used in GaussianAD. This design choice enables supervision using only BEV-level annotations (e.g., semantic maps and agent states), eliminating the need for dense 3D occupancy labels and significantly improving computational efficiency. 3) Additionally, GaussianFusion incorporates a dual-branch feature fusion design tailored for the planning-centric task.

\section{Method}
The objective of E2E autonomous driving is to predict the future trajectory of the ego vehicle directly from raw sensor inputs. Formally, given multi-view images $\mathcal{I}=\{\mathbf{I}_{i} \in \mathbb{R}^{3 \times H \times W} \mid i=1, \ldots, N\}$, LiDAR points $P \in \mathbb{R}^{n \times 3}$ and transformation matrices $\mathcal{T}=\{\mathbf{T}_{i} \in \mathbb{R}^{3 \times 3} \mid i=1, \ldots, N\}$, the goal is to predict the ego vehicle trajectory $\tau=\{\left(x_{t}, y_{t}\right)\}_{t=1}^{T}$, where $N$, $(H, W)$, $n$, $(x_t, y_t)$ and $T$ denotes the number of views, the image resolution, the number of points, the waypoint coordinate at time $t$ and the planning horizon, respectively.

\begin{figure*}[t]
    \centering
    \includegraphics[width=0.9\textwidth]{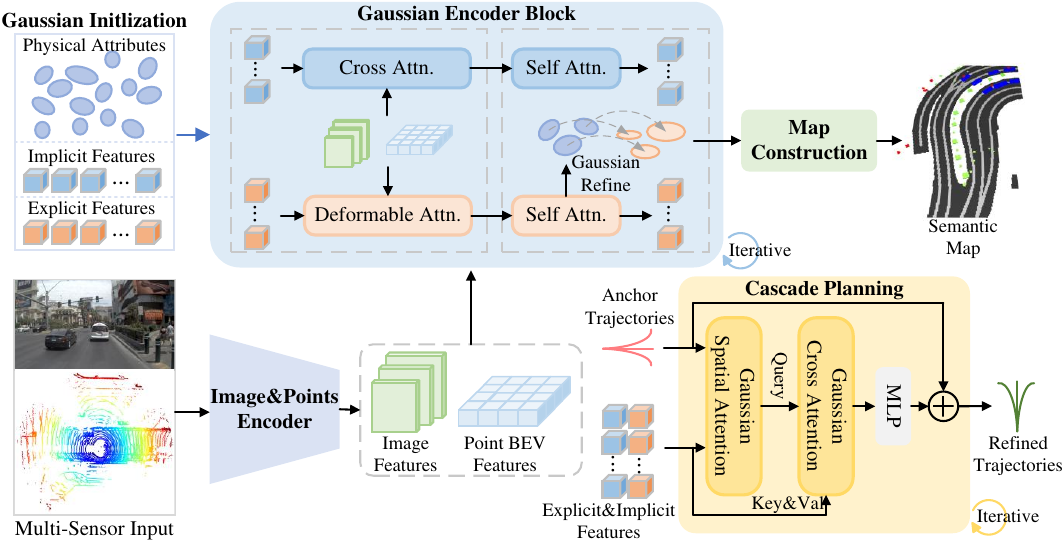}
    \caption{\textbf{The overall framework of our GaussianFusion.} Given raw multi-sensor data as input, GaussianFusion first extracts image and point features using a backbone network. It then initializes a set of Gaussians, which are iteratively refined through Gaussian encoder blocks. Finally, the refined Gaussians are used to construct the semantic map and to iteratively adjust the anchor trajectories. }
    \label{fig2}
\end{figure*}

The overview of our GaussianFusion is illustrated in Figure~\ref{fig2}. It can be divided into three phases as follows: (1) Gaussian initialization (Sec.~\ref{gaussian init}); (2) Gaussian encoder: Gaussians from multi-sensor (Sec.~\ref{gaussian encoder}); (3) Gaussian decoder: Gaussians to scene reasoning (Sec.~\ref{gaussian decoder}).

\subsection{Gaussian Initialization}
\label{gaussian init}

Given that autonomous vehicles primarily operate on planar surfaces, 2D Gaussians suffice to model the traffic scene. Specifically, we randomly generate a set of 2D Gaussians $\mathcal{G} = \{\mathbf{G}_i \mid i = 1, \ldots, P\}$, where $P$ denotes the predefined number of Gaussians. Each Gaussian is characterized by both physical attributes and hidden features. The physical attributes include mean $\mathbf{m} \in \mathbb{R}^{2}$, scale $\mathbf{s} \in \mathbb{R}^{2}$, rotation $\mathbf{r} \in \mathbb{R}^{2}$, and semantic logits $\mathbf{c} \in \mathbb{R}^{C}$, where $C$ is the number of semantic categories. The rotation $\mathbf{r}$ is represented using its sine and cosine components. The hidden features consist of both explicit and implicit features, which are then refined through two separate branches. Further details regarding the explicit and implicit features are provided in Sec.~\ref{gaussian encoder}.


\subsection{Gaussian Encoder: Gaussians from Multi-Sensor Fusion }
\label{gaussian encoder}
To learn meaningful Gaussian representations, we first utilize two independent backbones to extract multi-scale features from images and LiDAR points. These features are then used to iteratively refine both the physical attributes and hidden features of Gaussians. Each iteration consists of a point cross-attention module, an image cross-attention module, a Gaussian self-attention module, and a refinement module. The hidden features of each Gaussian are divided into explicit and implicit components, each serving distinct roles. Explicit features are derived from local regions of multi-sensor inputs via explicit geometric transformations and are responsible for updating the physical properties of Gaussians. In contrast, implicit features interact with global multi-sensor features without relying on geometric transformations, and are used exclusively for trajectory planning.

\noindent \textbf{Point Cross-Attention: Gaussians from Points. }
A point cross-attention (PCA) module is employed to extract information from point features. Specifically, for each Gaussian $\mathbf{G} = \{\mathbf{m}, \mathbf{s}, \mathbf{r}, \mathbf{c}, \mathbf{f}^{\text{exp}}, \mathbf{f}^{\text{imp}}\}$, we generate a set of query points $\mathcal{Q} = \{(x_i, y_i) \mid i = 1, \ldots, n_q\}$, where $(x_i,y_i)$ denotes the location of the $i$-th query point, and $n_q$ is the total number of queries. Following \cite{huang2024gaussianformer}, the query set includes both fixed and learnable points: the fixed queries are distributed around each Gaussian based on its covariance matrix, while the learnable queries are constrained within the interior of the Gaussian. Given multi-scale point feature maps $\mathcal{M}^{p} = \{\mathbf{M}^{p}_i \in \mathbb{R}^{d \times H_i^p \times W_i^p} \mid i = 1, \ldots, n_s\}$, where $(H_i^p, W_i^p)$ denotes the resolution of the $i$-th scale feature map and $n_s$ is the number of scales, we apply a deformable attention layer \cite{deform_atnn} to aggregate information from these features and update the explicit features of Gaussians.

\begin{equation}
    \begin{aligned}
        \mathbf{f}^{exp\dagger} = \sum_{i=1}^{n_q} \operatorname{DeAttn}\left(\mathbf{f}^{exp}, \mathcal{Q}[i], \mathcal{M}^{p}\right),
    \end{aligned}
    \label{Eq:eq1}
\end{equation}

\noindent where $\mathbf{f}^{exp\dagger}$ denotes the updated explicit features using multi-scale point features, $\mathcal{Q}[i]$ indicates the $i$-th point in $\mathcal{Q}$, and $\operatorname{DeAttn}(\cdot)$ represents the deformable attention. For the implicit features, we use a vanilla cross-attention \cite{cross-attention} to build interactions with the last scale point features $\mathbf{F}^p_{n_s}$:

\begin{equation}
    \begin{aligned}
        \mathbf{f}^{imp\dagger} =  \operatorname{CrossAttn}\left(\mathbf{f}^{imp}, \mathbf{M}^{p}_{n_s}\right),
    \end{aligned}
    \label{Eq:eq2}
\end{equation}

\noindent where $\mathbf{f}^{\text{imp}\dagger}$ denotes the updated implicit features, and $\operatorname{CrossAttn}$ refers to the cross-attention layer. For brevity, we omit the residual connection and feed-forward network (FFN) components in the formulation. The resulting updated Gaussian is represented as $\mathbf{G}^\dagger = \{\mathbf{m}, \mathbf{s}, \mathbf{r}, \mathbf{c}, \mathbf{f}^{\text{exp}\dagger}, \mathbf{f}^{\text{imp}\dagger}\}$. Note that, for clarity, we illustrate the process using a single Gaussian as an example.

\noindent \textbf{Image Cross-Attention: Gaussians from Images. }
To incorporate visual information from multi-view images, we employ an Image Cross-Attention (ICA) module. Similar to the PCA module, ICA generates both fixed and learnable query points for each Gaussian. However, these queries additionally incorporate height information to enable projection into the image plane. Specifically, we first generate 2D query points $\mathcal{Q}_{2d} = \{(x_i, y_i) \mid i = 1, \ldots, n_q\}$, identical to those used in the PCA module. For each 2D query location, we then uniformly sample $n_p$ pillar points along the vertical axis. The bottom of each pillar is fixed at $z = z_{\min}$, while the top is parameterized by a learnable variable $z_g \in [z_{\min}, z_{\max}]$, where $z_{\min}$ and $z_{\max}$ define the vertical bounds of the traffic scene. This results in a set of 3D query points $\mathcal{Q}_{3d} = \{(x_i, y_i, z_i) \mid i = 1, \ldots, n_q \times n_p\}$. Given the multi-scale image feature maps $\mathcal{M}^{I} = \{\mathbf{M}^{I}_i \in \mathbb{R}^{d \times N \times H_i^I \times W_i^I} \mid i = 1, \ldots, n_s\}$ extracted by the image backbone, where $(H_i^I, W_i^I)$ denote the resolution of the $i$-th scale feature map and $N$ is the number of camera views, the explicit and implicit features of Gaussians are computed as follows:

\begin{equation}
    \begin{aligned}
        \mathbf{f}^{exp\ddagger} = \sum_{i=1}^{n_q \times n_p} \operatorname{DeAttn}\left(\mathbf{f}^{exp\dagger}, \mathcal{Q}_{3d}\left[i\right], \mathcal{M}^{I}\right), \hspace{1em}
        \mathbf{f}^{imp\ddagger} =  \operatorname{CrossAttn}\left(\mathbf{f}^{imp\dagger}, \mathbf{M}^{I}_{n_s}\right),
    \end{aligned}
    \label{Eq:eq3}
\end{equation}

\noindent where $\mathcal{Q}_{3d}[i]$ denotes the $i$-th 3D query point in $\mathcal{Q}_{3d}$. Following the PCA module, we obtain the updated Gaussian representation $\mathbf{G}^\ddagger = \{\mathbf{m}, \mathbf{s}, \mathbf{r}, \mathbf{c}, \mathbf{f}^{\text{exp}\ddagger}, \mathbf{f}^{\text{imp}\ddagger}\}$.

\noindent \textbf{Gaussians Refinement Module. }
After aggregating information from multi-modal features, we further refine the Gaussian representations. Specifically, we employ two separate self-attention layers \cite{attention} to build interactions among all Gaussians--one for the explicit features and the other for the implicit features:

\begin{equation}
    \begin{aligned}
        \{\mathbf{f}_1^{exp\prime}, \ldots, \mathbf{f}_P^{exp\prime}\} &= \operatorname{SelfAttn}( \{\mathbf{f}_1^{exp\ddagger}, \ldots, \mathbf{f}_P^{exp\ddagger}\}, \{\mathbf{e}_1, \ldots, \mathbf{e}_P\}),\\
        \{\mathbf{f}_1^{imp\prime}, \ldots, \mathbf{f}_P^{imp\prime}\} &= \operatorname{SelfAttn}( \{\mathbf{f}_1^{imp\ddagger}, \ldots, \mathbf{f}_P^{imp\ddagger}\}, \{\mathbf{e}_1, \ldots, \mathbf{e}_P\}),\\
        \{\mathbf{e}_1, \ldots, \mathbf{e}_P\} &= \operatorname{PosEmbed}(\{\mathbf{m}_1,\ldots,\mathbf{m}_P\}),
    \end{aligned}
    \label{Eq:eq4}
\end{equation}

\noindent where $\mathbf{e}_i$ denotes the positional embedding of the $i$-th Gaussian, $\operatorname{SelfAtten}(\cdot)$ and $\operatorname{PosEmbed}(\cdot)$ refer to the self-attention and positional embedding layers \cite{dab-detr}, respectively. Subsequently, following \cite{huang2024gaussianformer}, a multi-layer perceptron (MLP) is employed to refine the physical attributes of Gaussians based on their explicit features:

\begin{equation}
    \begin{aligned}
        \mathbf{G}^{\prime} = \{\mathbf{m^\prime + m, \mathbf{s}^\prime, \mathbf{r}^\prime, \mathbf{c}^\prime}, \mathbf{f}^{exp\prime}, \mathbf{f}^{imp\prime}\},\hspace{1em}
        (\mathbf{m^\prime, \mathbf{s}^\prime, \mathbf{r}^\prime, \mathbf{c}^\prime}) = \operatorname{MLP}(\mathbf{f}^{exp\prime}).
    \end{aligned}
    \label{Eq:eq5}
\end{equation}

The Gaussian encoder described above is applied iteratively to refine the Gaussian representations. The final updated Gaussians are then passed to the Gaussian decoder, which performs downstream tasks such as mapping and planning.

\subsection{Gaussian Decoder: Gaussians to Scene Reasoning}
\label{gaussian decoder}
To effectively regulate the 2D Gaussians, we design a Gaussian decoder comprising two components: map construction and cascade planning. The map construction module explicitly reconstructs the traffic scene, providing backpropagation gradients that guide the Gaussian encoder in refining the physical attributes and explicit features. Following \cite{gaussianformerv2}, we implement this module using probabilistic Gaussian superposition; further details are provided in Appendix \ref{map_construction}. The cascade planning module generates trajectory predictions in a cascaded manner, where each subsequent output is refined based on the preceding one. In addition to leveraging the explicit features, it also incorporates the implicit features obtained from the Gaussian implicit fusion branch.

\noindent \textbf{Cascade Planning. }
We adopt an anchor-based planning strategy \cite{vadv2, li2024hydra, li2025hydra}, which constructs an anchor trajectory vocabulary based on the trajectory distribution observed in the dataset. Given the set of Gaussians obtained from the Gaussian encoder, we refine anchor trajectories $\mathcal{A}=\{\mathbf{A}_i \in \mathbb{R}^{T \times 2} \mid i = 1, \ldots, k\}$ in a cascaded manner, where $T$ denotes the planning horizon and the number of trajectory points. Taking a single anchor trajectory $\mathbf{A} \in \mathcal{A}$ as an example, we first compute the distance between each of its trajectory points and all Gaussians. For each point, we select its top-$m$ nearest Gaussians, forming a Gaussian subset $\mathcal{G}_A = \{\mathbf{G}_i \mid i = 1, \ldots, mT\}$. The anchor feature $\mathbf{F}_A$ is then obtained by querying this Gaussian set:

\begin{equation}
    \begin{aligned}
        &\mathbf{F}_A =\operatorname{CrossAttn}(\mathbf{F}_{query}, \mathbf{F}_{\mathcal{G}_A}), \\
        &\mathbf{F}_{query} = \operatorname{Embedding}(\mathbf{A}),\\ 
        &\mathbf{F}_{\mathcal{G}_A} = \{[\mathbf{f}^{exp}_i;\mathbf{f}^{imp}_i] \mid i=1,\ldots,mT\},
    \end{aligned}
    \label{Eq:eq6}
\end{equation}

\noindent where $\operatorname{CrossAttn}$ denotes a cross-attention layer, and $\operatorname{Embedding}$ represents an embedding layer that encodes the anchor trajectory $\mathbf{A}$ into an initial query feature $\mathbf{F}_{\text{query}}$. The terms $\mathbf{f}^{\text{exp}}_i$ and $\mathbf{f}^{\text{imp}}_i$ refer to the explicit and implicit features of the Gaussian $\mathbf{G}_i \in \mathcal{G}_A$, respectively, and $[\mathbf{f}^{\text{exp}}_i; \mathbf{f}^{\text{imp}}_i]$ denotes their concatenation. The traffic map and surrounding agents can be decoded from the most recent Gaussians $\mathcal{G}$, enabling $\mathcal{G}$ to serve as a comprehensive representation of the traffic scene. Therefore, we employ a cross-attention layer to build interactions between the anchor feature $\mathbf{F}_A$ and hidden features of $\mathcal{G}$. The refined trajectory $\tau=\{\left(x_{t}, y_{t}\right)\}_{t=1}^{T}$ is obtained as follows:

\begin{equation}
    \begin{aligned}
        & \tau = \operatorname{MLP}(\mathbf{F}_o) + \mathbf{A},\\
        &\mathbf{F}_o =\operatorname{CrossAttn}(\mathbf{F}_{A}, \mathbf{F}_{\mathcal{G}}), \\
        &\mathbf{F}_{\mathcal{G}} = \{[\mathbf{f}^{exp}_i;\mathbf{f}^{imp}_i] \mid i=1,\ldots,P\}.
    \end{aligned}
    \label{Eq:eq7}
\end{equation}

\noindent The trajectory is refined in a cascaded manner, where the output trajectory $\tau$ from the current stage is used as the anchor input for the subsequent stage, iteratively repeating the steps described in Eq.\ref{Eq:eq6} and Eq.~\ref{Eq:eq7} (referred to as Gaussian Spatial Attention and Gaussian Cross Attention in Figure~\ref{fig2}, respectively). We adopt the same trajectory loss function as proposed in \cite{diffusiondrive} to supervise the cascade refinement process.

\begin{table*}[t]
\centering
\resizebox{\linewidth}{!}{
\begin{tabular}{lccccccc}
\toprule
\textbf{Method} & \textbf{Input} & \textbf{Img. Backbone} & \textbf{NC} $\uparrow$ & \textbf{DAC} $\uparrow$ & \textbf{TTC} $\uparrow$ & \textbf{EP} $\uparrow$ & \textbf{PDMS} $\uparrow$ \\
\midrule
LTF & Camera & ResNet34 & 97.4 & 92.8 & 92.4 & 79.0 & 83.8 \\
TransFuser & C \& L & ResNet34 & 97.7 & 92.8 & 93.0 & 79.2 & 84.0 \\
GoalFlow & C \& L & ResNet34 & \textbf{98.3} & 93.3 & 94.8 & 79.8 & 85.7 \\
Hydra-MDP & C \& L & ResNet34 & \textbf{98.3} & 96.0 & 94.6 & 78.7 & 86.5 \\
Hydra-MDP++ & C & ResNet34 & 97.6 & 96.0 & 93.1 & 80.4 & 86.6 \\
ARTEMIS & C \& L & ResNet34 & \textbf{98.3} & 95.1 & 94.3 & 81.4 & 87.0 \\
DiffusionDrive & C \& L & ResNet34 & 98.2 & 96.2 & \textbf{94.7} & 82.2 & 88.1 \\
\rowcolor{cyan!20}
GaussianFusion (Ours) & C \& L & ResNet34 & \textbf{98.3} & \textbf{97.2} & 94.6 & \textbf{83.0} & \textbf{88.8} \\
\hline
GoalFlow & C \& L & V2-99 & 98.4 & 98.3 & 94.6 & 85.0 & 90.3 \\
Hydra-MDP-C & C & V2-99 & \textbf{98.7} & 98.2 & 95.0 & 86.5 & 91.0 \\
Hydra-MDP++ & C & V2-99 & 98.6 & \textbf{98.6} & 95.1 & 85.7 & 91.0 \\
\rowcolor{cyan!20}
GaussianFusion (Ours) & C \& L & V2-99 & \textbf{98.7} & 98.1 & \textbf{95.7} & \textbf{88.2} & \textbf{92.0} \\
\bottomrule
\end{tabular}}
\caption{\textbf{Performance on the Navtest Benchmark with Original Metrics.} Definitions of sub-metrics are provided in Appendix~\ref{metrics}. The best results of different backbones are highlighted in bold, separately.}
\label{tab:navsim_pdms}
\end{table*}

\begin{table*}[t]
\centering
\resizebox{\linewidth}{!}{
\begin{tabular}{lccccccc|c}
\toprule
\textbf{Method} & \textbf{NC} $\uparrow$ & \textbf{DAC} $\uparrow$ & \textbf{EP} $\uparrow$ & \textbf{TTC} $\uparrow$ & \textbf{DDC} $\uparrow$ & \textbf{LK} $\uparrow$ & \textbf{EC} $\uparrow$ & \textbf{EPDMS} $\uparrow$ \\
\midrule

TransFuser$^{*}$~\cite{transfuser} & 97.7 & 92.8 & 79.2 & 92.8 & 98.3 & 67.6 & 95.3 & 77.8 \\
VADv2$^{*}$~\cite{vadv2} & 97.3 & 91.7 & 77.6 & 92.7 & 98.2 & 66.0 & 97.4 & 76.6 \\
Hydra-MDP$^{*}$~\cite{li2024hydra} & 97.5 & 96.3 & 80.1 & 93.0 & 98.3 & 65.5 & 97.4 & 79.8 \\
Hydra-MDP++$^{*}$~\cite{li2025hydra} & 97.9 & 96.5 & 79.2 & 93.4 & 98.9 & 67.2 & 97.7 & 80.6 \\
ARTEMIS \cite{feng2025artemis} & \textbf{98.3} & 95.1 & 81.5 & 97.4 & 98.6 & 96.5 & 98.3 & 83.1 \\
DiffusionDrive \cite{diffusiondrive} & 98.2 & 96.2 & \textbf{87.6} & 97.3 & 98.6 & 97.0 & \textbf{98.4} & 84.0 \\
\rowcolor{cyan!20}
GaussianFusion (Ours) & \textbf{98.3} & \textbf{97.3}& 87.5 & \textbf{97.4} & \textbf{99.0} & \textbf{97.4} & 98.3 & \textbf{85.0} \\
\bottomrule
\end{tabular}}
\caption{\textbf{Performance on the NAVSIM \textit{navtest} split with extended metrics.} `*' represents that the results are sourced from Hydra-MDP++~\cite{li2025hydra}. The best results are highlighted in bold. }
\label{tab:navsim}
\end{table*}

\section{Experiments}

\subsection{Benchmark and Metric} 
We evaluate models on NAVSIM \cite{navsim} and Bench2Drive \cite{bench2drive} benchmarks. NAVSIM, built upon the OpenScene dataset \cite{openscene2023}, provides 120 hours of challenging driving data with high-resolution camera images and LiDAR inputs spanning up to 1.5 seconds. It filters out trivial driving scenarios to emphasize complex decision-making. We use the official Predictive Driver Model Score (PDMS) and its extended version, EPDMS \cite{li2025hydra}, as evaluation metrics. Bench2Drive \cite{bench2drive}, based on the CARLA simulator \cite{carla}, evaluates E2E autonomous driving across 220 routes that cover 44 interactive scenarios under diverse conditions. We adopt the CARLA Driving Score (DS) and multi-ability metrics for a fine-grained assessment. For more details about these metrics, please refer to Appendix~\ref{metrics}.

\section{Implementation Details}
For the NAVSIM benchmark, we use the NAVSIM $\textit{train}$ split for training. We utilize input from front, left-front, and right-front cameras, along with LiDAR point clouds. The camera images are cropped to a resolution of $448 \times 250$. For the Bench2Drive \cite{bench2drive} benchmark, we adopt only the front-view camera, which already provides a wide field of view. In this case, we use the resolution of $1024 \times 384$, which is consistent with the setting used in TransFuser++ \cite{hidden_datasets}. For the two benchmarks, LiDAR points are projected onto the BEV plane, following the approach in TransFuser \cite{transfuser}. In our main experiments, the number of Gaussians is set to 512, and each Gaussian feature has a dimensionality of 128. We adopt 4 GaussianEncoder blocks and 2 cascade planning blocks. The number of anchor trajectories is set to 20 following \cite{diffusiondrive}. It is worth noting that the map construction module in the GaussianDecoder is detached during inference for efficiency. Training is performed using the AdamW optimizer \cite{adamw}, with 50 epochs, a weight decay of $1 \times 10^{-4}$, and a maximum learning rate of $6 \times 10^{-4}$, which follows a cosine annealing schedule for learning rate decay. Hyper-parameter analysis is in Appendix~\ref{hyper_para_analysis}.

\begin{table*}[t]
    \centering
    \resizebox{\linewidth}{!}{
    \begin{tabular}{lcc|ccccc|c}
        \toprule
        \multirow{2}{*}{\textbf{Method}} & \multicolumn{2}{c|}{\textbf{Overall}$\uparrow$} & \multicolumn{5}{c|}{\textbf{Multi-Ability}$\uparrow$} & \\
        & \textbf{DS}  & \textbf{SR}  & \textbf{Merge}  & \textbf{Overtake} & \textbf{EBrake} & \textbf{GiveWay}  & \textbf{TSign}  & \textbf{Mean} $\uparrow$ \\
        \midrule
        PDM-Lite \cite{sima2024drivelm} & 97.0 & 92.3 & 88.8 & 93.3 & 98.3 & 90.0 & 93.7 & 92.8 \\
        \hline
        AD-MLP \cite{zhai2023rethinking} & 18.1 & 0.0 & 0.0 & 0.0 & 0.0 & 0.0 & 4.4 & 0.9 \\
        TCP \cite{tcp} & 40.7 & 15.0 & 16.2 & 20.0 & 20.0 & 10.0 & 7.0 & 14.6 \\
        VAD \cite{vad} & 42.4 & 15.0 & 8.1 & 24.4 & 18.6 & 20.0 & 19.2 & 18.1 \\
        UniAD \cite{uniad} & 45.8 & 16.4 & 14.1 & 17.8 & 21.7 & 10.0 & 14.2 & 15.6 \\
        ThinkTwice \cite{thinktwice} & 62.4 & 33.2 & 27.4 & 18.4 & 35.8 & \textbf{50.0} & 54.4 & 37.2 \\
        DriveTransformer \cite{jia2025drivetransformer} & 63.5 & 35.0 & 17.6 & 35.0 & 48.4 & 40.0 & 52.1 & 38.6\\
        DriveAdapter \cite{driveadapter} & 64.2 & 33.1 & 28.8 & 26.4 & 48.8 & \textbf{50.0} & 56.4 & 42.1 \\
        TF++$^{*}$ \cite{hidden_datasets} & 76.9$\pm$0.9 & 54.0$\pm$1.0 & \textbf{48.8}$\pm$2.2 & 37.6$\pm$7.0 & 64.2$\pm$8.4 & 50.0$\pm$0.0 & 59.7$\pm$6.1 & 52.0$\pm$3.0 \\
        \rowcolor{cyan!20}
        GaussianFusion (Ours) & \textbf{79.1}$\pm$1.1 & \textbf{54.4}$\pm$2.6 & 36.6$\pm$3.3 & \textbf{64.4}$\pm$2.3 & 66.5$\pm$6.4 & \textbf{53.3}$\pm$5.8 & \textbf{60.8}$\pm$6.4 & \textbf{56.3}$\pm$3.7 \\
        \bottomrule
    \end{tabular}}
    \caption{\textbf{Performance on the Bench2Drive benchmark.} `SR', `EBrake', and `TSign' denote the success rate, emergency braking, and traffic sign compliance, respectively. PDM-Lite is a rule-based planner that can access privileged information from the CARLA simulator. `*' denotes the implementation of the same backbone and training dataset setting as our method. }
\label{tab:b2d}
\end{table*}

\begin{table*}[t]
\centering
\resizebox{\linewidth}{!}{
\begin{tabular}{cccc|c|cccc|c}
\toprule
\multirow{2}{*}{\textbf{\textit{\shortstack{Gaussian \\ Exp. Fusion}}}} & \multirow{2}{*}{\textbf{\textit{\shortstack{Gaussian \\Imp. Fusion}}}} & \multirow{2}{*}{\textbf{\textit{\shortstack{Cascade \\Planning}}}} & \multirow{2}{*}{\textbf{\textit{\shortstack{Agent \\Pred.}}}} & \multirow{2}{*}{\textbf{Param.}} & \multirow{2}{*}{\textbf{DAC$\uparrow$}} & \multirow{2}{*}{\textbf{TTC$\uparrow$}} & \multirow{2}{*}{\textbf{EP$\uparrow$}} & \multirow{2}{*}{\textbf{LK$\uparrow$}} & \multirow{2}{*}{\textbf{EPDMS$\uparrow$}} \\
&&&&&&&&&\\
\midrule
\ding{55} & \ding{55} & \ding{55} & \ding{55} & 55.7M & 94.1 & 96.9 & 87.5 & 96.2 & 81.8 \\
\ding{51} & \ding{55} & \ding{55} & \ding{55} & 49.6M & 96.5 & 97.1 & \textbf{87.8} & 97.0 & 84.2 \\
\ding{51} & \ding{51} & \ding{55} & \ding{55} & 51.6M & 96.6 & 97.3 & \textbf{87.8} & 97.3 & 84.5 \\
\ding{51} & \ding{55} & \ding{51} & \ding{55} & 53.7M & 97.0 & 97.2 & 87.4 & 97.2 & 84.4 \\
\rowcolor{cyan!20}
\ding{51} & \ding{51} & \ding{51} & \ding{55} & 55.8M & \textbf{97.3} & \textbf{97.4} & 87.5 & \textbf{97.4} & \textbf{85.0} \\
\ding{51} & \ding{51} & \ding{51} & \ding{51} & 59.5M & 96.2 & 97.3 & 87.6 & 97.0 & 83.8 \\
\bottomrule
\end{tabular}}
\caption{\textbf{Ablation study} of different architecture configurations. We utilize TransFuser \cite{transfuser} without the agent prediction head as the baseline model (the first row setting). }
\label{tab:ablation}
\end{table*}

\subsection{Comparison with State-of-the-Art Methods}
\noindent\textbf{Results on NAVSIM.}
We benchmark GaussianFusion against leading state-of-the-art (SOTA) methods on the NAVSIM \textit{navtest} split. We employ ResNet34 \cite{resnet} and V2-99 \cite{V299} as the image backbone of GaussianFusion, respectively. As shown in Table~\ref{tab:navsim_pdms}, our method significantly surpasses previous methods, particularly on critical sub-metrics like DAC and Ego Vehicle Progress (EP) under the ResNet34 setting. Our enhanced GaussianFusion, equipped with V2-99, further improves performance across all metrics—achieving state-of-the-art results on the Navtest benchmark. This demonstrates the scalability of our framework with stronger backbones. Additionally, we evaluate performance under the EPDMS metric. Note that EPDMS poses a stricter challenge than PDMS by incorporating more nuanced driving criteria. As shown in Table~\ref{tab:navsim}, our approach achieves 85.0 EPDMS, outperforming recent strong baselines such as \cite{diffusiondrive} and \cite{feng2025artemis} by 1.0 and 1.9 points, respectively. A closer look reveals that most gains come from the Drivable Area Compliance (DAC) and Lane Keeping (LK) sub-metrics, indicating that GaussianFusion enables more stable and context-aware behavior in complex environments. These results consistently affirm the robustness and effectiveness of our method across multiple evaluation protocols.


\noindent\textbf{Results on Bench2Drive.}
We further conduct experiments on the closed-loop benchmark, Bench2Drive, to compare our method with existing SOTA E2E methods. Due to high performance variability in the CARLA simulator, we report results across three different random seeds. As shown in Table~\ref{tab:b2d}, our method, GaussianFusion, achieves the best overall performance (79.4 DS), outperforming all learning-based baselines. It shows balanced strength across diverse tasks, with notable gains in overtaking and traffic sign compliance. Compared to the rule-based privileged method PDM-Lite, our method still falls short to some extent, indicating that there remains substantial room for improvement in E2E autonomous driving methods.


\subsection{Ablation Study}
\noindent\textbf{Effect of Different Components.}
To understand the impact of each design choice in GaussianFusion, we perform a controlled ablation study by incrementally adding Gaussian Explicit Fusion (\textbf{\textit{Gaussian Exp. Fusion}}), Gaussian Implicit Fusion (\textbf{\textit{Gaussian Imp. Fusion}}), the Cascade Planning head (\textbf{\textit{Cascade Planning}}), and the Agent Prediction head (\textbf{\textit{Agent Pred.}}) to the TransFuser baseline \cite{transfuser}. The results are shown in Table~\ref{tab:ablation}. 
Introducing \textbf{\textit{Gaussian Exp. Fusion}} leads to a substantial +2.4 gain in EPDMS, while also reducing the parameter count. Adding \textbf{\textit{Gaussian Imp. Fusion}} further improves performance to 84.5 EPDMS, with only a slight increase in parameters. Separately, incorporating \textbf{\textit{Cascade Planning}} with only the explicit fusion also improves trajectory prediction, increasing LK and EPDMS compared to using explicit fusion alone. The best overall performance is achieved when combining dual-branch fusion and \textbf{\textit{Cascade Planning}}, reaching an EPDMS of 85.0. These findings confirm that our architectural components offer strong performance gains without significantly increasing model complexity. Finally, we observe that incorporating the agent prediction head degrades performance. We attribute this to the agent prediction task failing to provide effective guidance for the Gaussian refinement process, instead introducing ambiguity—what we refer to as Gaussian confusion. Given that the semantic map already encodes sufficient agent-related information, we remove the agent prediction head from our final design.

\noindent\textbf{Different Multi-Sensor Fusion Methods.}
Table~\ref{tab:fusion_methods} presents a comprehensive comparison of various multi-sensor fusion methods in terms of model parameters, semantic map construction, trajectory planning, and inference latency. The latency is measured by an RTX3090. To ensure a fair comparison, all methods adopt the same backbone and task heads. 

Compared with the BEV fusion method, our Gaussian fusion achieves better performance (+0.04 mIoU and +0.7 EPDMS), while also significantly reducing computational cost: the FLOPs are 38.9\% lower and latency is reduced by 25.6\%, showing clear advantages in both accuracy and efficiency.
\begin{wrapfigure}[9]{l}{0.5\textwidth}
  \centering
        \resizebox{\linewidth}{!}{
	\begin{tabular}{c|ccccc}
		\toprule
            \multirow{2}{*}{\textbf{\shortstack{Fusion \\Method}}} & \multirow{2}{*}{\textbf{Param.}} & \multirow{2}{*}{\textbf{FLOPs (G)}} & \multirow{2}{*}{\textbf{mIoU$\uparrow$}} & \multirow{2}{*}{\textbf{EPDMS$\uparrow$}} & \multirow{2}{*}{\textbf{\shortstack{Latency$\downarrow$ \\ (ms)}}}\\
            &&&&&\\
            \midrule
            Flatten & 55.7M & 27.4 & 0.50 & 81.8 & \textbf{18}\\
            BEV  & 51.9M & 58.4 & 0.51 & 83.8 & 43\\
            \rowcolor{cyan!15}
            Gaussian & 51.6M & 35.7 & \textbf{0.55} & \textbf{84.5} & 32\\
            \bottomrule
	\end{tabular}
        }
        \captionsetup{type=table}
	\caption{\textbf{Comparison of fusion methods.} }
        \label{tab:fusion_methods}
\end{wrapfigure} 
Compared with the flatten (TransFuser-style) fusion, which is a very lightweight approach, our method exhibits only a 30.3\% increase in FLOPs, while achieving much stronger performance (+0.05 mIoU and +2.7 EPDMS). However, our method's latency is 77.8\% higher, which we believe is largely due to the differences in operator-level optimization. This could be further mitigated through engineering improvements in deployment.

\begin{figure*}[t]
    \centering
    \includegraphics[width=1.0\textwidth]{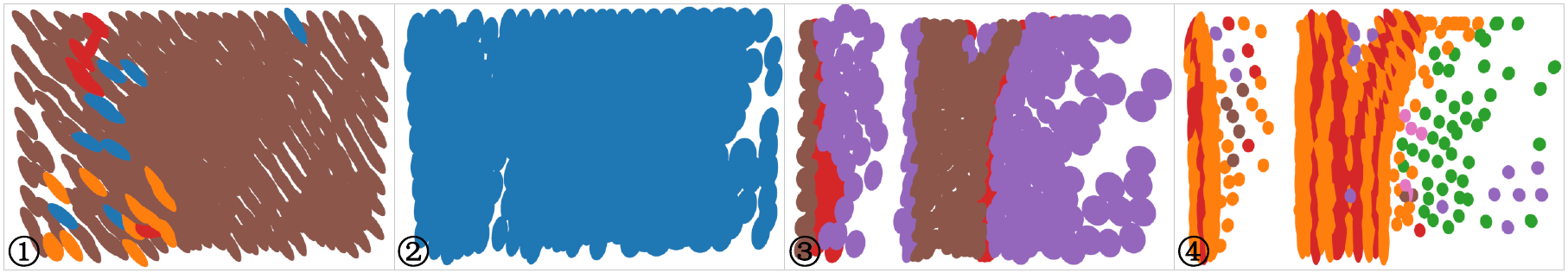}
    \caption{\textbf{Visualization of Gaussians} during the refinement process. Gaussians with different semantics are shown in different colors.}
    \label{fig3}
\end{figure*}

\begin{figure*}[t]
    \centering
    \includegraphics[width=1.0\textwidth]{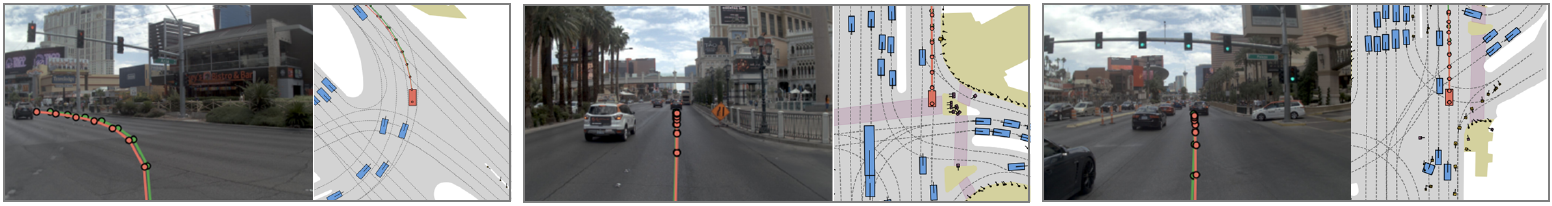}
    \caption{\textbf{Visualization of predicted and ground-truth trajectories}, shown in red and green, respectively.}
    \label{fig4}
\end{figure*}

\subsection{Qualitative Comparison}

To provide an intuitive understanding of the refinement process in the Gaussian encoder, we visualize the spatial distribution of Gaussians at different refinement stages, as shown in Figure~\ref{fig3}. At the initial stage, Gaussians are evenly dispersed throughout the scene. As refinement proceeds, they progressively converge toward foreground regions. This behavior highlights the advantages of the Gaussian representation, which offers a more compact and adaptable alternative to traditional dense BEV maps. More visualizations of Gaussians are shown in Figure~\ref{fig5} in the Appendix. We also present predicted ego-vehicle trajectories under a variety of traffic scenarios in Figure~\ref{fig4}. To qualitatively evaluate the prediction accuracy, we compare these trajectories against ground-truth data. In the leftmost scenario of Figure~\ref{fig4}, the vehicle makes an unprotected left turn without signal guidance—a challenging case. Our method still predicts a trajectory closely matching the ground truth. In addition, as shown in the two rightmost scenarios of Figure~\ref{fig4}, our method is capable of producing accurate trajectory plans even in dense traffic conditions, further demonstrating its robustness and reliability.

\section{Conclusion}
In this work, we propose GaussianFusion, a Gaussian-based multi-sensor fusion framework for E2E autonomous driving. By utilizing compact and flexible 2D Gaussian representations, our method balances spatial awareness with computational efficiency. A dual-branch fusion architecture captures both local details and global planning cues from multi-modal inputs, while a cascade planning module progressively refines trajectory predictions. Experiments on NAVSIM and Bench2Drive benchmarks show that GaussianFusion significantly improves planning performance with high efficiency. These results underscore the promise of Gaussian representations for efficient and interpretable sensor fusion in the E2E autonomous driving system. The limitation of GaussianFusion lies in its customized CUDA operations, which are not fully optimized. In future work, we plan to either further optimize these operations or replace them with operations of a well-established neural network library.

\noindent \textbf{Limitations.} We have not conducted quantitative evaluations on tracking small or fast-moving objects. Similarly, our method has not been assessed under sensor occlusion or noise, which are critical factors for real-world deployment. Moreover, our method depends on 3D bounding box and road topology labels to optimize the 2D Gaussians. While effective, such supervision can be expensive or unavailable in real-world scenarios.

\section*{Acknowledgments}
\noindent This work was supported in part by the the Guangxi Key R \& D Program under Grant No.GuikeAB24010324, and in part by the Guangdong Basic and Applied Basic Research Foundation under Grant 2025A1515011485.

\bibliographystyle{plainnat}
\bibliography{ref}


\clearpage
\appendix

\begin{figure*}[t]
    \centering
    \includegraphics[width=1.0\textwidth]{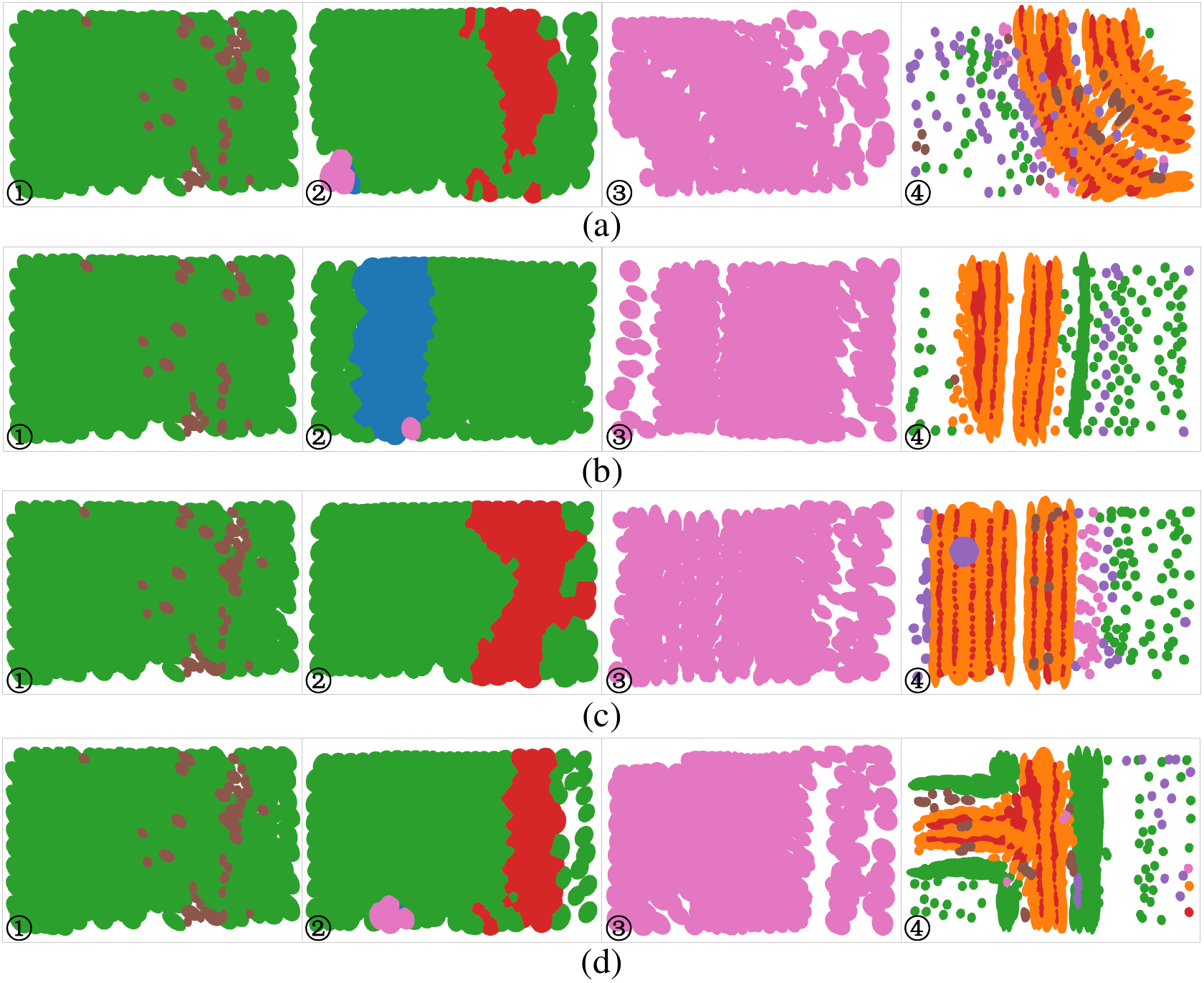}
    \caption{\textbf{Visualization of Gaussians} of different inputs during the refinement process. Gaussians with different semantics are shown in different colors.}
    \label{fig5}
\end{figure*}

\section{Map Construction}
\label{map_construction}
Following \cite{gaussianformerv2}, we utilize Gaussian probability superposition to construct the semantic map. Specifically, for a map pixel position $\mathbf{x} \in \mathbb{R}^2$, we first calculate its probability of falling in different Gaussians:

\begin{equation}
    \begin{aligned}
        \alpha(\mathbf{x} ; \mathbf{G}_i)=\exp \left(-\frac{1}{2}(\mathbf{x}-\mathbf{m}_i)^{\mathrm{T}} \boldsymbol{\Sigma}_i^{-1}(\mathbf{x}-\mathbf{m}_i)\right), \\
        \boldsymbol{\Sigma}_i=\mathbf{R S S}^{T} \mathbf{R}^{T}, \quad \mathbf{S}=\operatorname{diag}(\mathbf{s}_i), \quad \mathbf{R}=\mathrm{r}2\mathrm{m}(\mathbf{r}_i),
    \end{aligned}
    \label{Eq:eq8}
\end{equation}

\noindent where \(\mathbf{m}_i\), \(\mathbf{s}_i\), and \(\mathbf{r}_i\) denote the mean, scale, and rotation vectors of the Gaussian \(\mathbf{G}_i\), respectively. \(\boldsymbol{\Sigma}_i\) represents the covariance matrix of \(\mathbf{G}_i\); \(\operatorname{diag}(\cdot)\) denotes the operator that constructs a diagonal matrix from a given vector; and \(\operatorname{r2m}(\cdot)\) denotes the function that converts a rotation vector into its corresponding rotation matrix. Then the superposition probability can be calculated as:

\begin{equation}
    \begin{aligned}
        \mathbf{o}(\mathbf{x} ; \mathcal{G}) & =\sum_{i=1}^{P} p\left(\mathbf{G}_{i} \mid \mathbf{x}\right) \mathbf{c}_{i}^\prime=\frac{\sum_{i=1}^{P} p\left(\mathbf{x} \mid \mathbf{G}_{i}\right) a_{i} \mathbf{c}_{i}^\prime}{\sum_{j=1}^{P} p\left(\mathbf{x} \mid \mathbf{G}_{j}\right) a_{j}} \in \mathbb{R}^C, \\
        p\left(\mathbf{x} \mid \mathbf{G}_{i}\right) & =\frac{1}{(2 \pi)^{\frac{3}{2}}|\boldsymbol{\Sigma}_i|^{\frac{1}{2}}} \exp \left(-\frac{1}{2}(\mathbf{x}-\mathbf{m})^{\mathrm{T}} \boldsymbol{\Sigma}^{-1}_i(\mathbf{x}-\mathbf{m})\right),
        \end{aligned}
    \label{Eq:eq9}
\end{equation}

\noindent where $\mathbf{o}(\mathbf{x};\mathbf{G}_i)$ denotes the logits of different semantics on $\mathbf{x}$, $a_i$ represents the prior probability of the existence of $i$-th Gaussian which is predicted by the network, and $p(\mathbf{G}_i \mid \mathbf{x})$ indicates the posterior probability of the $i$-th Gaussian being present given the observation at $\mathbf{x}$. Based on these quantities, we can derive the probabilities for the background and foreground:

\begin{equation}
    \begin{aligned}
        p_\mathrm{b}(\mathbf{x}) &= \prod_{i=1}^{P} (1 - \alpha(\mathbf{x}; \mathbf{G}_i)) \in \mathbb{R}^1,\\
        p_\mathrm{f}(\mathbf{x}) &= (1 - p_b(\mathbf{x})) \operatorname{Softmax}(\mathbf{o}(\mathbf{x};\mathcal{G})) \in \mathbb{R}^C,
    \end{aligned}
    \label{Eq:eq10}
\end{equation}

\noindent where $p_\mathrm{b}(\mathbf{x})$ and $p_\mathrm{f}(\mathbf{x})$ denote the background and foreground probabilities of pixel $\mathbf{x}$, $\operatorname{Softmax}(\cdot)$ represents the Softmax function. Intuitively, for a background pixel $\mathbf{x}_\mathrm{b}$, the $p_\mathrm{b}(\mathbf{x}_\mathrm{b})$ will decrease with Gaussians gather on $\mathbf{x}_\mathrm{b}$ because $p_\mathrm{b}(\mathbf{x}_\mathrm{b}) \le (1- \alpha(\mathbf{x}_\mathrm{b};\mathbf{G}_i))$ holds for any Gaussian. Conversely, for a foreground pixel $\mathbf{x}_f$, the $p_\mathrm{f}(\mathbf{x}_\mathrm{f})$ will increase with Gaussians gather on $\mathbf{x}_\mathrm{f}$ due to $1 - p_\mathrm{b}(\mathbf{x}_\mathrm{f}) \ge \alpha(\mathbf{x}_\mathrm{f};\mathbf{G}_i)$ holds for any Gaussian. Therefore, Eq.~\ref{Eq:eq10} will encourage Gaussians to move to foreground pixels, as demonstrated in Figure~\ref{fig3}. Given background and foreground predictions $[ p_\mathrm{b}; p_\mathrm{f}] \in \mathbb{R}^{C+1}$ and ground-truth bev map, we employ cross entropy loss $L_{ce}$ and the lovasz-softmax \cite{lovasz} loss $L_{lov}$ as the map construction loss functions.

\begin{figure}
    \begin{minipage}[h]{.49\linewidth}
	\centering
        \resizebox{\linewidth}{!}{
	\begin{tabular}{c|cccc|c}
		\toprule
            \multirow{2}{*}{\textbf{\shortstack{Gaussian \\Numbers}}} & \multirow{2}{*}{\textbf{DAC$\uparrow$}} & \multirow{2}{*}{\textbf{TTC$\uparrow$}} & \multirow{2}{*}{\textbf{EP$\uparrow$}} & \multirow{2}{*}{\textbf{LK$\uparrow$}} & \multirow{2}{*}{\textbf{EPDMS$\uparrow$}} \\
            &&&&&\\
            \midrule
            256 & 96.6 & 97.3 & \textbf{87.6} & 97.1 & 84.1 \\
            \rowcolor{cyan!15}
            512 & \textbf{97.3} & 97.4 & 87.5 & 97.4 & \textbf{85.0}\\
            768 & 97.2 & \textbf{97.5} & 87.4 & \textbf{97.5} & 84.8 \\
            \bottomrule
	\end{tabular}
        }
        \captionsetup{type=table}
	\caption{\textbf{Impact of the number of Gaussians.} }
        \label{tab:gas_num}
    \end{minipage}
    \hfill
    \begin{minipage}[h]{.49\linewidth}
	\centering
        \resizebox{\linewidth}{!}{
	\begin{tabular}{c|cccc|c}
		\toprule
            \multirow{2}{*}{\textbf{\shortstack{Encoder \\Blocks}}} & \multirow{2}{*}{\textbf{DAC$\uparrow$}} & \multirow{2}{*}{\textbf{TTC$\uparrow$}} & \multirow{2}{*}{\textbf{EP$\uparrow$}} & \multirow{2}{*}{\textbf{LK$\uparrow$}} & \multirow{2}{*}{\textbf{EPDMS$\uparrow$}} \\
            &&&&&\\
            \midrule
            3 & 97.1 & \textbf{97.5} & 87.4 & \textbf{97.5} & 84.6 \\ 
            \rowcolor{cyan!15}
            4 & \textbf{97.3} & 97.4 & 87.5 & 97.4 & \textbf{85.0}\\
            5 & 97.1 & \textbf{97.5} & \textbf{87.7} & 97.2 & 84.7 \\
            \bottomrule
	\end{tabular}
        }
        \captionsetup{type=table}
	\caption{\textbf{Effect of the number of encoder blocks.} }
        \label{tab:encoder_num}
    \end{minipage}    
\end{figure}

\begin{wrapfigure}{l}{0.5\textwidth}
  \centering
    \resizebox{\linewidth}{!}{
	\begin{tabular}{c|cccc|c}
		\toprule
            \multirow{2}{*}{\textbf{\shortstack{Cascade \\Stages}}} & \multirow{2}{*}{\textbf{DAC$\uparrow$}} & \multirow{2}{*}{\textbf{TTC$\uparrow$}} & \multirow{2}{*}{\textbf{EP$\uparrow$}} & \multirow{2}{*}{\textbf{LK$\uparrow$}} & \multirow{2}{*}{\textbf{EPDMS$\uparrow$}} \\
            &&&&&\\
            \midrule
            1 & 96.6 & \textbf{97.4} & 87.4 & 97.1 & 84.3 \\
            \rowcolor{cyan!15}
            2 & \textbf{97.3} & \textbf{97.4} & 87.5 & \textbf{97.4} & \textbf{85.0}\\
            3 & 97.0 & \textbf{97.4} & \textbf{87.8} & 97.3 & 84.7 \\
            \bottomrule
	\end{tabular}
    }
    \captionsetup{type=table}
    \caption{\textbf{Results of the stage of cascade planning.} }
    \label{tab:cas_num}
\end{wrapfigure} 

\section{Metrics}
\label{metrics}
For the NAVSIM \cite{navsim} benchmark, we adopt the official Predictive Driver Model Score (PDMS) and its extended version, Extended PDMS \cite{li2025hydra} (EPDMS). These metrics are designed to bridge the gap between open-loop planning benchmarks and closed-loop simulation evaluations. PDMS comprises multiple sub-metrics, including No-Collision (NC), Drivable Area Compliance (DAC), Time-to-Collision (TTC), Comfort (C), and Ego Vehicle Progress (EP). EPDMS extends this framework by incorporating additional criteria: Lane Keeping (LK), Extended Comfort (EC), Driving Direction Compliance (DDC), Traffic Light Compliance (TLC), and False-Positive Penalty Filtering, thereby offering a more comprehensive assessment of driving behavior.

For the Bench2Drive \cite{bench2drive} benchmark, we employ the CARLA Driving Score (DS), which jointly considers route completion and penalties for traffic infractions. In addition to DS, we also use multi-ability metrics that target specific driving skills of the autonomous driving system, such as merging, overtaking, yielding, traffic sign compliance, and emergency braking.

\section{Hyper-Parameter Analysis}
\label{hyper_para_analysis}
In this section, we determine the optimal value for the number of Gaussians, the number of encoder blocks, and the stage number of cascade planning through experiments on the NAVSIM dataset. Table~\ref{tab:gas_num} shows the impact of the number of Gaussians. When increasing the number of Gaussians from 256 to 512, the performance is improved significantly. However, further enlarging the number of Gaussians does not bring gains. Table~\ref{tab:encoder_num} presents the effect of the number of Gaussian encoder blocks. Setting the number of encoder blocks to 4 is the best choice. Since we only supervise the output of the last encoder block, too deep may affect performance. Table~\ref{tab:cas_num} illustrates the performance of different stage number of cascade planning. We observe that 2 cascade stages are enough to generate accurate trajectories. 

\begin{figure*}[t]
    \centering
    \includegraphics[width=1.0\textwidth]{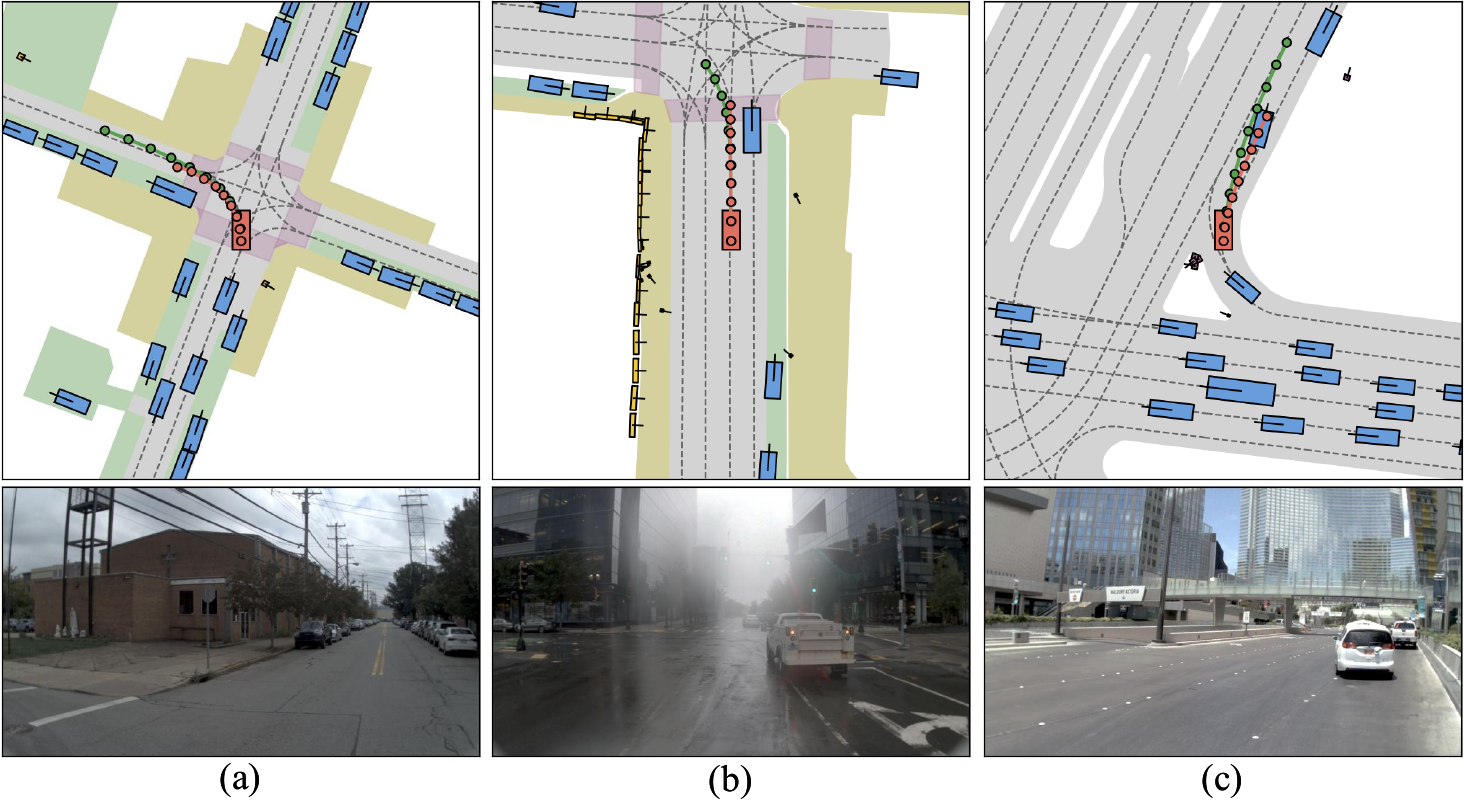}
    \caption{\textbf{Failure mode} of GaussianFusion in some scenarios.}
    \label{fig6}
\end{figure*}

\section{Failure Mode}
\label{failure_mode}
To better understand the limitations of GaussianFusion in complex real-world scenarios, we visualize typical failure cases in Figure~\ref{fig6}. These cases reveal distinct behavioral biases in the learned policy, reflecting the implicit priors encoded in the training dataset and the conservative nature of GaussianFusion.

In Figure~\ref{fig6}(a), the model demonstrates overly conservative behavior when executing turning maneuvers at intersections. Compared with human experts who perform smooth and anticipatory turns, GaussianFusion tends to delay or narrow the turning trajectory. We hypothesize that this is attributed to the data distribution, where the majority of demonstrated driving behaviors follow cautious and robust strategies, leading the model to learn a risk-averse policy in ambiguous topological regions.

Figure~\ref{fig6}(b) illustrates model behavior under adverse weather conditions such as fog. In low-visibility scenarios, the model prioritizes stability by choosing a slow and straight driving strategy rather than making proactive turning decisions. This behavior suggests that Gaussian primitives lose discriminative power in degraded sensor conditions, causing the planner to rely on the safest longitudinal control instead of predictive planning.

Figure~\ref{fig6}(c) reflects the model’s tendency to maintain its lane and follow a leading vehicle that is changing lanes, instead of performing an overtaking maneuver. This indicates a bias toward conservative following behavior in highly dynamic multi-agent scenarios. The model may underestimate the feasibility or safety margin of overtaking due to uncertainty in surrounding vehicle intentions, resulting in a reactive rather than assertive driving strategy.

Overall, these failure cases highlight that GaussianFusion inherits conservative driving priors from the dataset and tends to favor low-risk behaviors when facing uncertainty. While such conservatism improves safety in most situations, it may limit driving efficiency or adaptability in complex environments.

\end{document}